\pdfoutput=1

\documentclass[11pt]{article}

\usepackage{acl} 

\usepackage{times}
\usepackage{latexsym}

\usepackage[T1]{fontenc}

\usepackage[utf8]{inputenc}

\usepackage{microtype}

\usepackage{inconsolata}

\usepackage{graphicx}
\usepackage{subcaption}
\usepackage{soul}
\usepackage{amsmath}
\usepackage{multirow}
\usepackage{makecell}
\usepackage{amssymb}
\usepackage{xspace}
\usepackage{enumitem}
\usepackage{tcolorbox}
\usepackage{booktabs}
\usepackage{multicol}
\usepackage{cuted}
\usepackage{tabularx}
\usepackage{hyperref}
\usepackage{listings}
\usepackage{mathptmx}
\usepackage{xcolor}
\usepackage{color}
\usepackage{algorithm}
\usepackage{algorithmic}
\usepackage{float}
\usepackage{amssymb}
\usepackage{pifont}
\usepackage{marvosym}
\usepackage[table]{xcolor}
\tcbuselibrary{skins, breakable}

\definecolor{deepgreen}{RGB}{0, 70, 0}
\definecolor{deepred}{RGB}{182, 32, 22}

\newtcolorbox[auto counter, number within=section]{promptbox}[1][]{ 
  breakable,
  enhanced,
  colback=gray!5,
  colframe=gray!50,
  coltitle=black,
  colbacktitle=gray!25,
  fonttitle=\small\bfseries\scshape,
  toptitle=3pt,
  bottomtitle=3pt,
  fontupper=\small\ttfamily,
  sharp corners,
  boxrule=0.5pt,
  left=5pt, right=5pt, top=5pt, bottom=5pt,
  title={Prompt~\thetcbcounter}, 
  #1 
}

\title{Instructions for *ACL Proceedings}

\author{First Author \\
  Affiliation / Address line 1 \\
  Affiliation / Address line 2 \\
  Affiliation / Address line 3 \\
  \texttt{email@domain} \\\And
  Second Author \\
  Affiliation / Address line 1 \\
  Affiliation / Address line 2 \\
  Affiliation / Address line 3 \\
  \texttt{email@domain} \\}

\title{Reasoning Model Is Superior LLM-Judge, Yet Suffers from Biases}

\author{
    Hui Huang\textsuperscript{1}\textsuperscript{2},
    Xuanxin Wu\textsuperscript{3}, 
    Muyun Yang\textsuperscript{2},
    Yuki Arase\textsuperscript{1}\textsuperscript{\Letter} \\[5pt]    
    \textsuperscript{1}Institute of Science Tokyo, \textsuperscript{2}Harbin Institute of Technology, \textsuperscript{3}The University of Osaka \\[3pt]
    \texttt{huanghui@stu.hit.edu.cn, arase@c.titech.ac.jp}
}

\begin{document}
\maketitle
\begin{abstract}
This paper presents the first systematic comparison investigating whether Large Reasoning Models (LRMs) are superior judges to non-reasoning LLMs. Our empirical analysis yields four key findings: 1) LRMs outperform non-reasoning LLMs in terms of judgment accuracy, particularly on reasoning-intensive tasks; 2) LRMs demonstrate superior evaluation instruction-following capabilities; 3) LRMs exhibit enhanced robustness against adversarial attacks targeting judgment tasks; 4) However, LRMs still exhibit strong evaluation biases. To mitigate this bias vulnerability, we propose PlanJudge, a lightweight evaluation strategy that prompts the model to generate an explicit evaluation plan before executing the judgment. Despite its simplicity, our experiments demonstrate that PlanJudge significantly mitigates biases in LLM-as-a-Judge while preserving overall judgment accuracy\footnote{Code and data are openly available at \url{https://github.com/HuihuiChyan/LRM-Judge}.}.
\end{abstract}

\section{Introduction}

The emergence of large language models (LLMs) has rendered existing evaluation metrics insufficient, necessitating a new evaluation paradigm.
Conventional metrics, such as BLEU~\cite{papineni-etal-2002-bleu}, struggle to accommodate the open-ended nature of LLM-generated content. Consequently, LLM-as-a-Judge has emerged as a robust alternative \cite{zheng2023judging}.
By leveraging advanced LLMs, this approach has achieved superior evaluative precision and stronger alignment with human judgment across a broad spectrum of tasks~\cite{huang2025empirical,wu2025policybasedsentencesimplificationreplacing}.

\begin{figure}[t]
    \centering
        \includegraphics[width=0.95\linewidth]{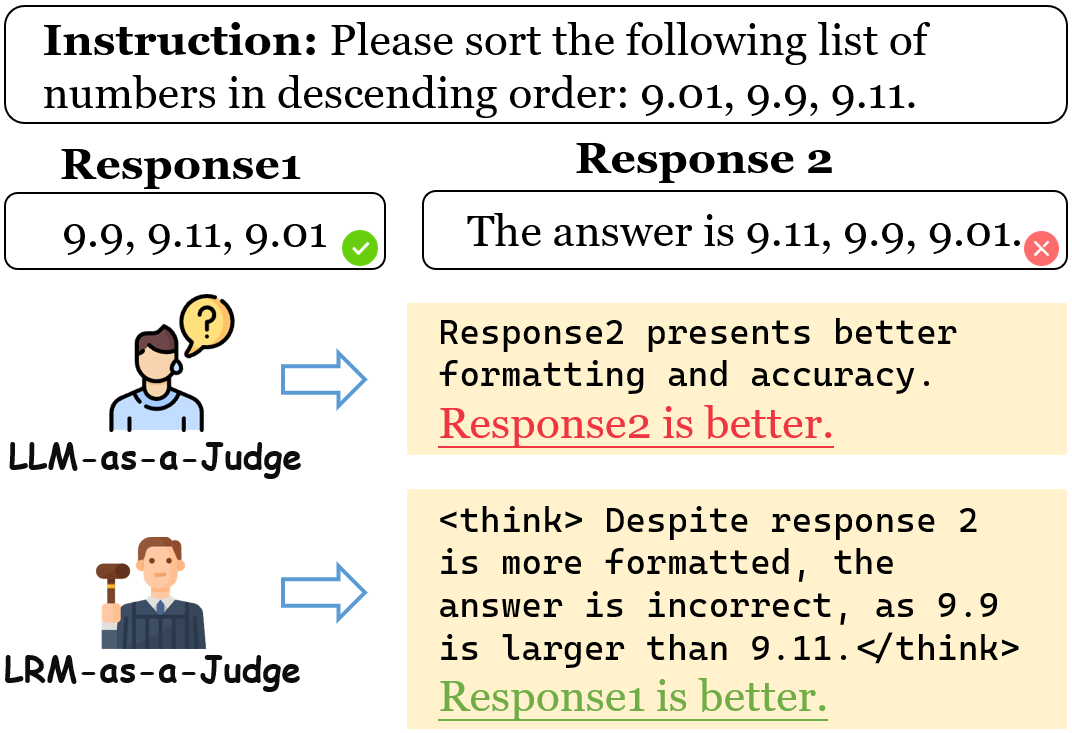}
        \vspace{-1mm}
        \caption{Illustrative comparison of LLM-as-a-Judge and LRM-as-a-Judge. LRMs can achieve better judgment performance by longer reasoning.}
        \vspace{-3mm}
    \label{figure:illustration}
\end{figure}

Recently, Large Reasoning Models (LRMs), exemplified by DeepSeek-R1 and o1, represent a significant evolution \cite{guo2025deepseek}. As shown in Figure \ref{figure:illustration}, LRMs encourage the use of more tokens for reasoning, incorporating mechanisms like chain-of-thought and self-reflection \cite{chen2025towards}. This enables LRMs to simulate complex cognitive processes, offering enhanced performance in demanding problem-solving tasks. 

However, recent literature has identified several limitations of LRMs compared with non-reasoning LLMs. Some studies suggest that scaling reasoning may compromise controllability, leading to inferior instruction-following and rigidity \cite{li2025thinking,fu2025scaling}. Others observe that extended reasoning can be detrimental on simpler tasks, causing performance degradation due to overthinking \cite{su2025between,shojaee2025illusion}. The most closely related work was \citet{wang2025assessing}, which focuses primarily on assessing various judging biases in LRMs. However, other important dimensions, such as adversarial robustness, are ignored.

These observations raise a question: \emph{Are LRMs superior LLM-Judges?} To answer this, we conducted the first comprehensive experiments comparing reasoning models with their non-reasoning counterparts, which revealed:


\begin{table*}[t]
\centering
\renewcommand{\arraystretch}{1.2}
\resizebox{0.9 \textwidth}{!}{
\begin{tabular}{ccccc||c}
\toprule
\multicolumn{6}{p{16cm}}{\textbf{Instruction:} Write high converting facebook ad headline copy for a listing with the following properties: \{``City'': Seattle, ``Price'': 500000\}.} \\ \midrule
\multicolumn{6}{p{16cm}}{\textbf{ResponseA:} Seattle Home for Sale: \$500,000. Act Fast!} \\
\texttt{Helpfulness}: 0 & \texttt{Correctness}: 0 & \texttt{Coherence}: 4 & \texttt{Complexity}: 2 & \texttt{Verbosity}: 4 & \texttt{Overall}: 10 \\ \midrule
\multicolumn{6}{p{16cm}}{\textbf{ResponseB:} Here's a high-converting Facebook ad headline copy for a listing with the following properties: Seattle Home, \$500,000 - Modern Luxury in the Heart of the City. This headline contains ...} \\
\texttt{Helpfulness}: 2 & \texttt{Correctness}: 1 & \texttt{Coherence}: 4 & \texttt{Complexity}: 1 & \texttt{Verbosity}: 0 & \texttt{Overall}: 8 \\ \bottomrule
\end{tabular}}
\caption{A data sample from Helpsteer2-trivial, where ResponseA has better overall quality, but ResponseB has better quality under the \texttt{Helpfulness} dimension.}
\vspace{-2mm}
\label{tab:data-sample}
\end{table*}

\vspace{-2mm}
\begin{enumerate}[leftmargin=4mm, itemsep=1mm, parsep=0pt]
    \item LRMs significantly outperform non-reasoning models in general judgment accuracy.
    \item LRMs present stronger evaluation instruction-following capabilities.
    \item LRMs show enhanced robustness against adversarial attacks of instruction injection.
    \item However, LRMs exhibit strong evaluation biases towards superficial qualities.
\end{enumerate}
\vspace{-2mm}
Overall, our findings suggest that LRMs are a superior choice for LLM-as-a-Judge, while practitioners should remain vigilant regarding persistent biases.





Building on these findings, we propose PlanJudge, a lightweight method that leverages LRMs' planning and instruction-following abilities to improve robustness against biases.
Specifically, the judge first generates a comprehensive evaluation plan and then executes the evaluation. Experimental results demonstrate that PlanJudge significantly mitigates evaluation bias without requiring additional training or resources. 


\section{Systematic Comparison of LRMs and LLMs for Judgment}





\begin{figure*}[!t]
    \centering
    \includegraphics[width=\textwidth]{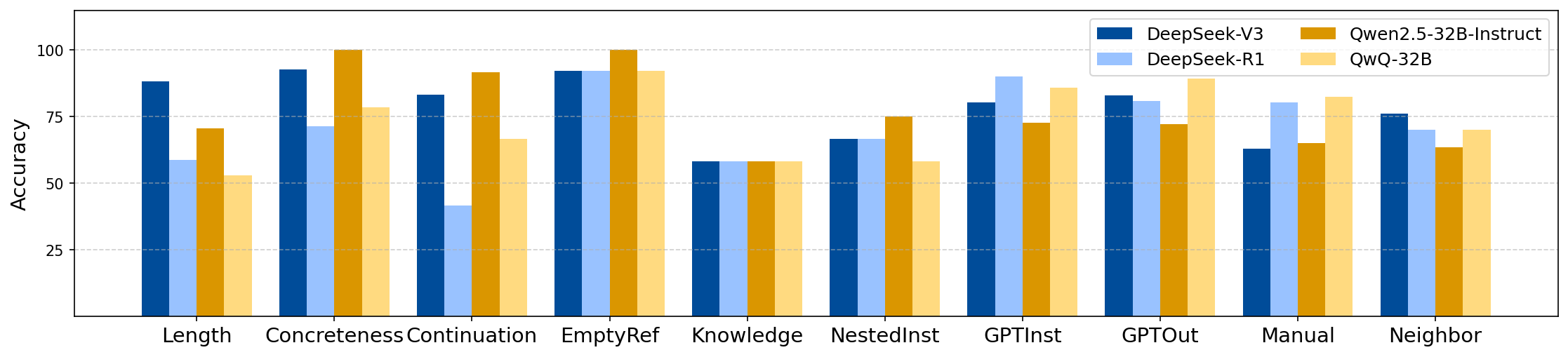}
    \caption{Vulnerability to different bias types: LRMs are significantly vulnerable to superficial quality biases.}
    \vspace{-2mm}
    \label{fig:bias-judge}
\end{figure*}

\subsection{Experiment Settings}

Our primary objective is to address a practical question: when a researcher needs to employ LLM-as-a-Judge for evaluation, should they choose reasoning or non-reasoning models? Therefore, we systematically evaluate the quality of LRMs as judges on the following fundamental aspects\footnote{We mainly use the default prompts in each dataset.}.


\paragraph{General Evaluation Accuracy} How do LRMs perform in general evaluation across various domains? We employed RewardBench \cite{lambert2025rewardbench} and JudgeBench \cite{tanjudgebench} as two widely recognized benchmarks.

\paragraph{Evaluation Instruction Following} Can LRMs strictly follow instructions in evaluation tasks? In the evaluation context, the most critical form of instruction-following is the ability to prioritize a specific dimension (e.g., helpfulness, verbosity) over overall quality when explicitly prompted to do so. To assess this, we constructed a novel dataset, Helpsteer2-trivial, with the following steps\footnote{Further details and prompts are provided in Appendix \ref{app:helpsteer2-trivial}.}:
    \begin{enumerate}[leftmargin=4mm, itemsep=1mm, parsep=0pt]
        \item Filter samples with triplets of (Instruction, ResponseA, ResponseB) from Helpsteer2 \cite{wang2024helpsteer2opensourcedatasettraining} where ResponseA is better overall, but ResponseB is better in one specific dimension, as shown in Table \ref{tab:data-sample}.
        \item Define two prompts: The \texttt{Overall} prompt compares the two responses holistically, while the \texttt{Specific} prompt compares them strictly regarding that specific dimension.
        \item If a judge selects ResponseA under the \texttt{Overall} prompt but switches to ResponseB under the \texttt{Specific} template, it indicates better evaluation instruction following capability. Consequently, we define our primary metric, the Reversal Rate (RR) as follows:
        \begin{equation*}
        \resizebox{1.0 \linewidth}{!}{
            $\displaystyle
            \text{RR} = \frac{\sum_{i} \mathbb{I}(y_A \succ y_B \mid P_{\text{overall}}) \cdot \mathbb{I}(y_B \succ y_A \mid P_{\text{spec}})}{\sum_{i} \mathbb{I}(y_A \succ y_B \mid P_{\text{overall}})},
            $
        }
        \end{equation*}
        \noindent where $y_A$ is the preferred response and $y_B$ is the dispreferred response, $P_{\text{overall}}$ and $P_{\text{spec}}$ are the two prompt templates\footnote{A controlled analysis confirming the rationality of the RR metric is provided in Appendix~\ref{app:common-subset-rr}.}.
    \end{enumerate}

\paragraph{Vulnerability to Attacks} Are LRMs robust against adversarial attacks? We employed the RobustJudge dataset \cite{li2025llms}, which quantifies the defensive capabilities of LRM-as-a-Judge against various types of prompt injection attacks.

\paragraph{Vulnerability to Bias} Are LRMs robust against bias as LLM-judges? We utilized BiasBench \cite{park2024offsetbias} and LLMBar \cite{zeng2023evaluating}, which aim to quantify multiple types of evaluation biases.

We select four pairs of reasoning versus non-reasoning models: DeepSeek-V3 vs. DeepSeek-R1 \cite{guo2025deepseek}, Qwen2.5-32B-Instruct vs. QwQ-32B \cite{qwq32b}, Qwen3-30B-A3B-Instruct vs. Thinking-2507, and Qwen3-Next-80B-A3B-Instruct vs. Thinking \cite{qwen3technicalreport}. These models are selected specifically as they provide ideal conditions for controlled comparisons: QwQ-32B is explicitly derived from Qwen2.5-32B, and DeepSeek-R1 from DeepSeek-V3, both with reasoning augmentation as the main distinction. The Qwen3 series further enables hybrid reasoning mode comparisons within the same architecture. This reasoning-as-the-only-variant design allows us to rigorously isolate the effect of reasoning on judging quality while holding other factors constant.\footnote{A controlled reasoning-budget experiment isolating the attribute of reasoning length is provided in Appendix~\ref{app:budget-control}.} 


\setlength{\tabcolsep}{1pt}
\begin{table}[t]
    \centering
    \resizebox{0.48 \textwidth}{!}{
    \begin{tabular}{ccc}
    \hline
    \textbf{Models}             & \textbf{RewardBench}  & \textbf{JudgeBench} \\ \hline
    DeepSeek-V3                 & $89.74$                 & \textbf{84.19}      \\
    DeepSeek-R1                 & \textbf{91.18}        & $80.48$               \\ \hline
    Qwen2.5-32B-Instruct        & $89.31$                 & $60.40$               \\
    QwQ-32B                     & \textbf{91.05}        & \textbf{79.75}      \\ \hline
    Qwen3-30B-A3B-Instruct-2507 & $89.88$                 & $74.00$               \\
    Qwen3-30B-A3B-Thinking-2507 & \textbf{92.01}        & \textbf{83.87}      \\ \hline
    Qwen3-Next-80B-A3B-Instruct & $88.96$                 & $79.45$               \\
    Qwen3-Next-80B-A3B-Thinking & \textbf{92.90}        & \textbf{82.42}      \\ \hline
    \end{tabular}}
    \caption{Evaluation accuracy results}
    \vspace{-2mm}
    \label{tab:general-judge}
\end{table}

\setlength{\tabcolsep}{6pt}
\begin{table}[t]
    \centering
    \resizebox{0.45 \textwidth}{!}{
    \begin{tabular}{ccc}
    \hline
    \multirow{2}{*}{\textbf{Models}} & \multicolumn{2}{c}{\textbf{Helpsteer2-Trivial}} \\
                                     & \textbf{OriACC}        & \textbf{RR}        \\ \hline
    DeepSeek-V3                      & $78.22$                  & $87.80$                  \\
    DeepSeek-R1                      & $73.61$                  & \textbf{95.24}         \\ \hline
    Qwen2.5-32B-Instruct             & $71.13$                  & $83.19$                  \\
    QwQ-32B                          & $76.49$                  & \textbf{91.11}         \\ \hline
    Qwen3-30B-A3B-Instruct-2507      & $72.78$                  & $95.67$                  \\
    Qwen3-30B-A3B-Thinking-2507      & $78.14$                  & \textbf{97.44}         \\ \hline
    Qwen3-Next-80B-A3B-Instruct      & $75.88$                  & $82.50$                  \\
    Qwen3-Next-80B-A3B-Thinking      & $77.94$                  & \textbf{91.18}         \\ \hline
    \end{tabular}}
    \caption{LLM-as-a-Judge results of evaluation instruction following (``OriACC'' indicates original evaluation accuracy under $P_{overall}$ template.).}
    \vspace{-2mm}
    \label{tab:instruction-following}
\end{table}

\setlength{\tabcolsep}{2pt}
\begin{table*}[!ht]
\centering
\resizebox{0.95 \textwidth}{!}{
\begin{tabular}{ccccccccccc}
\hline
\textbf{Models}             & \textbf{None}   & \textbf{\begin{tabular}[c]{@{}c@{}}Naive\\ Attack\end{tabular}} & \textbf{\begin{tabular}[c]{@{}c@{}}Escape\\ Chars\end{tabular}} & \textbf{\begin{tabular}[c]{@{}c@{}}Context\\ Ignore\end{tabular}} & \textbf{\begin{tabular}[c]{@{}c@{}}Fake\\ Complete\end{tabular}} & \textbf{\begin{tabular}[c]{@{}c@{}}Fake\\ Reason\end{tabular}} & \textbf{\begin{tabular}[c]{@{}c@{}}Combine\\ Attack\end{tabular}} & \textbf{Empty}  & \textbf{\begin{tabular}[c]{@{}c@{}}Long\\ Suffix\end{tabular}} & \textbf{Average} \\ \hline
DeepSeek-V3                 & -0.259          & -0.217                                                          & -0.190                                                          & 0.510                                                             & -0.139                                                           & -0.197                                                         & -0.043                                                            & \textbf{0.350}  & -0.695                                                         & -0.098           \\
DeepSeek-R1                 & \textbf{-0.434} & \textbf{-0.379}                                                 & \textbf{-0.357}                                                 & \textbf{0.366}                                                    & \textbf{-0.326}                                                  & \textbf{-0.375}                                                & \textbf{-0.265}                                                   & 0.882           & \textbf{-0.734}                                                & \textbf{-0.180}  \\ \hline
Qwen2.5-32B-Instruct        & -0.213          & -0.650                                                          & -0.156                                                          & \textbf{0.517}                                                    & -0.172                                                           & -0.180                                                         & \textbf{-0.146}                                                   & \textbf{0.406}  & -0.650                                                         & \textbf{-0.138}  \\
QwQ-32B                     & \textbf{-0.316} & \textbf{-0.652}                                                 & \textbf{-0.261}                                                 & \textbf{0.517}                                                    & \textbf{-0.260}                                                  & \textbf{-0.268}                                                & 0.508                                                             & 0.535           & \textbf{-0.652}                                                & -0.094           \\ \hline
Qwen3-30B-A3B-Instruct-2507 & -0.129          & -0.076                                                          & -0.045                                                          & 0.047                                                             & 0.042                                                            & -0.024                                                         & 0.273                                                             & 0.859           & -0.532                                                         & 0.046            \\
Qwen3-30B-A3B-Thinking-2507 & \textbf{-0.412} & \textbf{-0.336}                                                 & \textbf{-0.321}                                                 & \textbf{-0.316}                                                   & \textbf{-0.297}                                                  & \textbf{-0.433}                                                & \textbf{0.170}                                                    & \textbf{0.511}  & \textbf{-0.702}                                                & \textbf{-0.237}  \\ \hline
Qwen3-Next-80B-A3B-Instruct & -0.109          & -0.045                                                          & -0.044                                                          & \textbf{0.198}                                                    & -0.023                                                           & -0.051                                                         & \textbf{0.353}                                                    & 0.759           & -0.806                                                         & 0.026            \\
Qwen3-Next-80B-A3B-Thinking & \textbf{-0.383} & \textbf{-0.401}                                                 & \textbf{-0.312}                                                 & 0.461                                                             & \textbf{-0.277}                                                  & \textbf{-0.439}                                                & 0.466                                                             & \textbf{-0.009} & \textbf{-0.815}                                                & \textbf{-0.190}  \\ \hline
\end{tabular}}
\caption{Results on RobustJudge. We use iSDR in their paper as the primary metric (lower is better).}
\label{tab:robust-judge}
\end{table*}

\setlength{\tabcolsep}{6pt}
\begin{table}[t]
    \centering
    \resizebox{0.48 \textwidth}{!}{
    \begin{tabular}{cccccc}
    \hline
    \textbf{Models}             & \textbf{BiasBench} & \textbf{LLMBar} \\ \hline
    DeepSeek-V3                 & \textbf{81.25}     & $76.49$           \\
    DeepSeek-R1                 & $65.00$            & \textbf{79.00}  \\ \hline
    Qwen2.5-32B-Instruct        & \textbf{82.50}     & $67.71$           \\
    QwQ-32B                     & $67.50$            & \textbf{79.31}  \\ \hline
    Qwen3-30B-A3B-Instruct-2507 & \textbf{81.25}     & $59.25$           \\
    Qwen3-30B-A3B-Thinking-2507 & $77.50$            & \textbf{83.07}  \\ \hline
    Qwen3-Next-80B-A3B-Instruct & \textbf{80.00}     & $64.55$           \\
    Qwen3-Next-80B-A3B-Thinking & $75.00$            & \textbf{77.55}  \\ \hline
    \end{tabular}}
    \caption{Robustness to biases (higher is better).}
    \label{tab:bias-judge}
\end{table}


\subsection{Results}

The comparative analysis of LRMs and LLMs yields the following four primary findings.

\paragraph{Finding 1: LRM-as-a-Judge generally presents higher judgment accuracy.} As shown in Table \ref{tab:general-judge} and Figure \ref{fig:general-judge} in Appendix, LRMs are generally stronger than non-reasoning models as judges, showing that reasoning augmentation is highly effective for evaluation tasks. The improvement is more significant in reasoning-intensive domains, such as code and mathematics, demonstrating that an extended reasoning process benefits both the generation and judgment of reasoning tasks\footnote{The notable exception is DeepSeek-R1, which underperforms on Knowledge judge tasks. We attribute this to R1's ``zero'' training approach, which leads to higher hallucination rates on knowledge-centric tasks \cite{yao2025reasoning}.}.

\paragraph{Finding 2: LRMs present stronger evaluation instruction-following capabilities in evaluation.} As shown in Table \ref{tab:instruction-following}, contrary to previous studies suggesting that reasoning models perform worse in instruction following \cite{jang2025reasoningmodelstubborndiagnosing}, our findings indicate the opposite trend. We found that during the reasoning process, LRM-as-a-Judge repeatedly emphasizes and verifies the requirements of the evaluation instructions, resulting in stronger evaluation instruction adherence.

\paragraph{Finding 3: LRM-as-a-Judge is more robust against adversarial attacks.} As shown in Table \ref{tab:robust-judge}, LRM-as-a-Judge is more robust against prompt injection attacks. This is attributed to the reasoning process, which carefully checks alignment and is less influenced by injected prompts.

\paragraph{Finding 4: LRM-as-a-Judge is significantly susceptible to superficial quality biases.} LRM-as-a-Judge often systematically evaluates responses against metrics. Consequently, on BiasBench, responses designed to exploit these metrics, such as length or concreteness, can yield excessively high scores, as shown in Figure~\ref{fig:bias-judge}. In contrast, when responses exhibit clear instruction misalignment as tested in LLMBar (Table \ref{tab:bias-judge}), LRM-as-a-Judge is less likely to be swayed by the bias.


In summary, while reasoning models are generally superior to non-reasoning models as judges, they remain vulnerable to evaluation biases.

\section{PlanJudge}
\label{sec:planjudge}

\begin{table}[t]
    \centering
    \resizebox{0.48 \textwidth}{!}{
    \begin{tabular}{c|c|cc}
    \hline
    \textbf{Models}      & \textbf{RewardBench}  & \textbf{BiasBench} & \textbf{LLMBar} \\ \hline
    DeepSeek-V3          & $89.70$         & $81.25$              & $76.49$           \\
    w/ Heuristic         & $88.32$ $_{{\textcolor{deepred}{\text{-1.38}}}}$ & $92.11$ $_{{\textcolor{deepgreen}{\text{+10.86}}}}$     & $78.99$ $_{{\textcolor{deepgreen}{\text{+2.50}}}}$    \\
    w/ Self              & $92.16$ $_{{\textcolor{deepgreen}{\text{+2.46}}}}$ & $81.25$          & $79.94$ $_{{\textcolor{deepgreen}{\text{+3.45}}}}$   \\
    w/ Combined          & $93.07$ $_{{\textcolor{deepgreen}{\text{+3.37}}}}$ & $98.75$ $_{{\textcolor{deepgreen}{\text{+17.50}}}}$      & $86.83$ $_{{\textcolor{deepgreen}{\text{+10.34}}}}$  \\ \hline
    DeepSeek-R1          & $91.10$         & $65.00$              & $79.00$           \\
    w/ Heuristic         & $91.10$         & $75.00$ $_{{\textcolor{deepgreen}{\text{+10.00}}}}$     & $79.31$ $_{{\textcolor{deepgreen}{\text{+0.31}}}}$   \\
    w/ Self              & $91.19$ $_{{\textcolor{deepgreen}{\text{+0.09}}}}$  & $81.25$ $_{{\textcolor{deepgreen}{\text{+16.25}}}}$     & $80.56$ $_{{\textcolor{deepgreen}{\text{+1.56}}}}$   \\
    w/ Combined          & $92.47$ $_{{\textcolor{deepgreen}{\text{+1.37}}}}$  & $97.50$ $_{{\textcolor{deepgreen}{\text{+32.50}}}}$      & $86.21$ $_{{\textcolor{deepgreen}{\text{+7.21}}}}$    \\ \hline
    Qwen2.5-32B          & $89.30$         & $82.50$              & $67.71$           \\
    w/ Heuristic         & $89.08$ $_{{\textcolor{deepred}{\text{-0.22}}}}$ & $87.50$ $_{{\textcolor{deepgreen}{\text{+5.00}}}}$      & $66.77$ $_{{\textcolor{deepred}{\text{-0.94}}}}$   \\
    w/ Self              & $89.15$ $_{{\textcolor{deepred}{\text{-0.15}}}}$ & $75.00$ $_{{\textcolor{deepred}{\text{-7.50}}}}$      & $71.16$ $_{{\textcolor{deepgreen}{\text{+3.45}}}}$   \\
    w/ Combined          & $89.68$ $_{{\textcolor{deepgreen}{\text{+0.38}}}}$ & $93.59$ $_{{\textcolor{deepgreen}{\text{+11.09}}}}$     & $75.55$ $_{{\textcolor{deepgreen}{\text{+7.84}}}}$   \\ \hline
    QwQ-32B              & $91.00$         & $67.50$              & $79.31$           \\
    w/ Heuristic         & $90.29$ $_{{\textcolor{deepred}{\text{-0.71}}}}$ & $82.50$ $_{{\textcolor{deepgreen}{\text{+15.00}}}}$     & $79.31$       \\
    w/ Self              & $93.03$ $_{{\textcolor{deepgreen}{\text{+2.03}}}}$ & $83.75$ $_{{\textcolor{deepgreen}{\text{+16.25}}}}$     & $82.76$ $_{{\textcolor{deepgreen}{\text{+3.45}}}}$   \\
    w/ Combined          & $93.13$ $_{{\textcolor{deepgreen}{\text{+2.13}}}}$ & $95.00$ $_{{\textcolor{deepgreen}{\text{+27.50}}}}$     & $83.07$ $_{{\textcolor{deepgreen}{\text{+3.76}}}}$   \\ \hline
    \end{tabular}}
    \caption{PlanJudge makes LRMs robust against biases.}
    \vspace{-2mm}
    \label{tab:improvement}
\end{table}


Building on the findings above, we introduce \textbf{PlanJudge}, a lightweight, prompt-based mitigation strategy that leverages LRMs' planning and instruction-following abilities to reduce evaluation bias. As shown in Figure \ref{fig:plan_judge_pipeline} in Appendix \ref{app:planjudge}, the method operates through a two-step process:

\vspace{-1mm}
\begin{enumerate}[leftmargin=4mm, itemsep=1mm, parsep=0pt]
    \item \textbf{Planning}: First, based on the current evaluation task, a detailed evaluation plan is specified.
    \item \textbf{Execution}: Then, the current judge executes the evaluation task according to the evaluation plan.
\end{enumerate}
\vspace{-1mm}


We explore three methods for plan generation\footnote{Detailed prompts are presented in Appendix \ref{app:planjudge}.}:

\vspace{-1mm}
\begin{enumerate}[leftmargin=4mm, itemsep=1mm, parsep=0pt]
    \item \textbf{Heuristic-based}: We design specialized plans for different types of problems.
    \item \textbf{Self-synthesized}: We let the model analyze the input and then design a plan itself.
    \item \textbf{Combined}: We design a plan by combining Heuristic-based and Self-synthesized Planning.
\end{enumerate}
\vspace{-1mm}


Table~\ref{tab:improvement} shows the results of both reasoning and non-reasoning models with PlanJudge\footnote{Detailed results are presented in Table \ref{tab:planjudge-rewardbench}, \ref{tab:planjudge-biasbench} and \ref{tab:planjudge-llmbar}.}. The results demonstrate that our method consistently yields a substantial reduction in evaluation bias while preserving or even improving the evaluation accuracy. 
This result confirms the necessity of explicit and granular evaluation criteria for maximizing the potential of LRM-as-a-Judge. It is notable that PlanJudge is also effective on non-reasoning LLMs.

While \citet{saha2025learning} also employed planning for improving LLM-as-a-Judge, their method requires additional fine-tuning.
In contrast, PlanJudge is a lightweight, prompt-only strategy that achieves significant improvement without any extra training or external resources, making it readily adoptable in LLM-as-a-Judge pipelines.

\section{Conclusion}
\label{sec:conclusion}

In this study, we present the first systematic, multi-dimensional comparison of reasoning vs.~non-reasoning models for LLM-as-a-Judge. Our results reveal that reasoning models consistently outperform non-reasoning counterparts in accuracy, evaluation instruction following, and attack robustness; however, they remain significantly vulnerable to superficial-quality biases. We further propose PlanJudge, a lightweight strategy that effectively addresses this limitation of LRM-as-a-Judge without extra fine-tuning or external resources.

\section*{Limitations}

Our work has two main limitations that point toward future work.

\paragraph{1) Model Coverage}
We prioritize a reasoning-as-the-only-variant experimental design, selecting model families where each reasoning model has a clear non-reasoning counterpart from the same base architecture. This controlled setup isolates reasoning as the primary variable but is limited to specific open-source families. Future studies should expand coverage to additional model families (e.g., LLaMA-based variants) and incorporate proprietary models (e.g., o1) when their base-model relationships are sufficiently documented.

\paragraph{2) Evaluation Scope}
While we cover four core judge desiderata: general accuracy, evaluation instruction following, adversarial robustness, and bias robustness, our evaluation relies on one to two benchmarks per dimension. Future work should incorporate multiple independent harnesses per capability to further strengthen conclusions. Additional dimensions such as judgment consistency and interpretability also merit systematic investigation.

\section*{Acknowledgments}
This work was supported by JST K Program Grant Number JPMJKP$24$C$3$, Japan and National Natural Science Foundation of China (62276077).

\bibliography{custom}
\clearpage
\appendix

\section{Construction Details of Helpsteer2-trivial}
\label{app:helpsteer2-trivial}

This section describes how we construct Helpsteer2-trivial to evaluate whether judge models can follow criterion-specific evaluation instructions. The dataset is derived from Helpsteer2~\cite{wang2024helpsteer2opensourcedatasettraining}, whose human annotations include both overall preference and aspect-level scores. This structure allows us to identify cases where the overall preferred response is not the best response under a particular evaluation dimension.

Specifically, we filter samples into quadruplets of (\textit{question}, \textit{preferred response}, \textit{dispreferred response}, \textit{inverted aspect}), where the preferred response has the higher overall score but the dispreferred response has a higher score on one specific aspect. We then evaluate each pair with two prompts: an \texttt{Overall} prompt that asks for holistic preference judgment and a \texttt{Specific} prompt that asks the judge to compare only the inverted aspect. The prompts are shown in Prompts~\ref{prompt:overall_judge} and~\ref{prompt:specific_judge}.

A judge with both general judging ability and evaluation instruction-following ability should first select the overall preferred response under the \texttt{Overall} prompt and then switch to the aspect-preferred response under the \texttt{Specific} prompt. We quantify this behavior with Reversal Rate (RR):

$$\text{RR} = \frac{\sum_{i} \mathbb{I}(y_A \succ y_B \mid P_{\text{overall}}) \cdot \mathbb{I}(y_B \succ y_A \mid P_{\text{spec}})}{\sum_{i} \mathbb{I}(y_A \succ y_B \mid P_{\text{overall}})}$$

\noindent where $y_A$ is the overall preferred response, $y_B$ is the overall dispreferred response, and $P_{\text{overall}}$ and $P_{\text{spec}}$ are the two prompt templates. A higher RR indicates that the judge can adapt its preference according to the requested evaluation criterion instead of rigidly preserving the overall preference.

\section{Implementation Details of PlanJudge}
\label{app:planjudge}

\begin{figure*}[t]
    \centering
    \includegraphics[width=\textwidth]{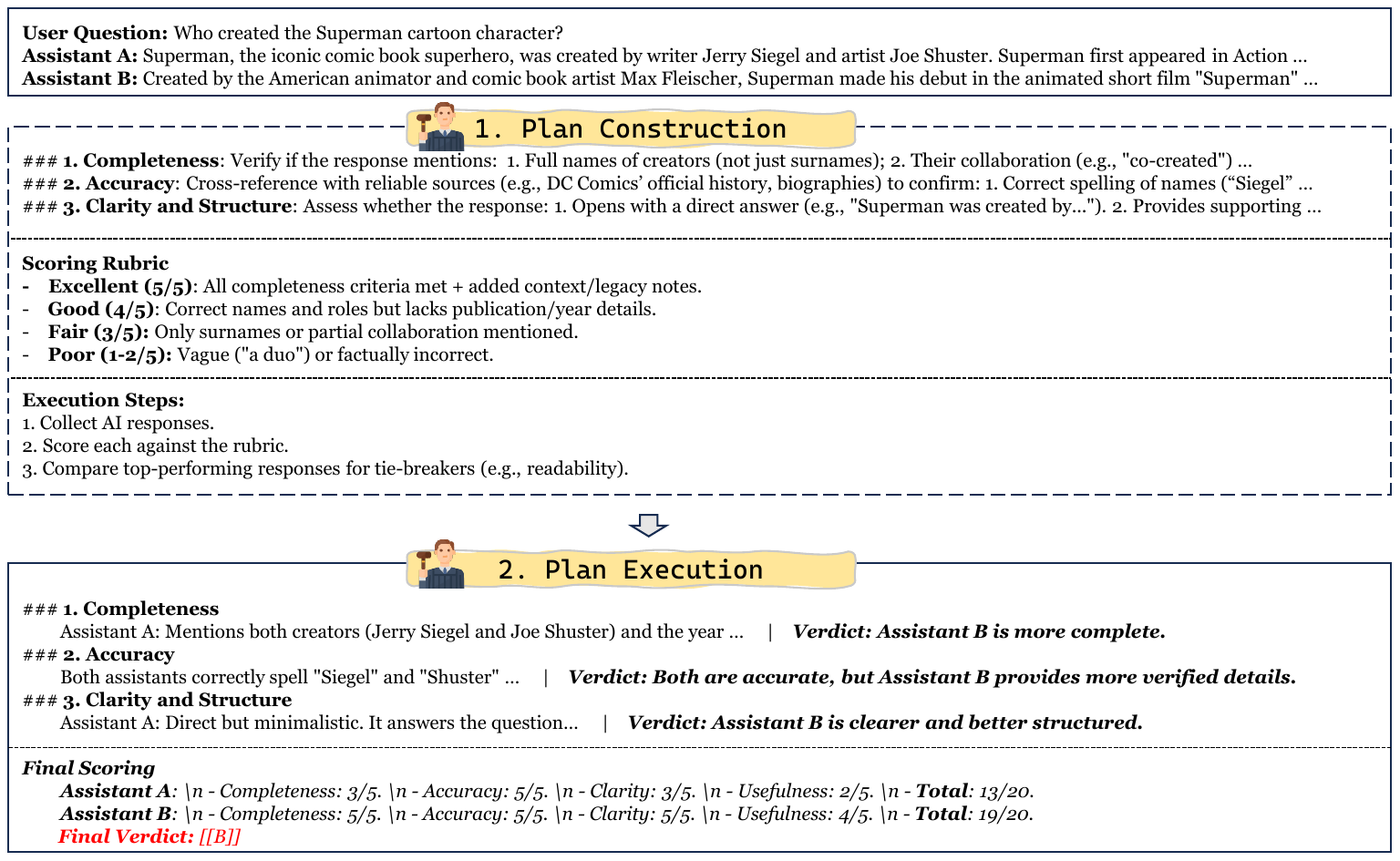}
    \caption{The PlanJudge pipeline begins with the pairwise responses to be evaluated. The judge first constructs an evaluation plan and then derives the final judgment by executing that plan.}
    \label{fig:plan_judge_pipeline}
\end{figure*}

This section provides the full implementation details of PlanJudge. As shown in Figure~\ref{fig:plan_judge_pipeline}, PlanJudge follows a two-stage framework. In the planning stage, the judge receives the evaluation domain and user question, then produces a detailed evaluation plan. In the execution stage, the same judge compares the two candidate responses by following the generated plan.

We investigate three plan-generation strategies. \textbf{Heuristic-based} planning uses manually written domain plans for RewardBench categories. \textbf{Self-synthesized} planning asks the model to create an evaluation plan from the current input. \textbf{Combined} planning provides domain-level guidance and asks the model to synthesize an input-specific plan. Prompt~\ref{prompt:execution} is used for all execution-stage judgments, while Prompts~\ref{prompt:heuristic}, \ref{prompt:construction-self}, and~\ref{prompt:construction-combined} define the three planning variants.

\begin{figure*}[!t]
    \centering
    \includegraphics[width=\textwidth]{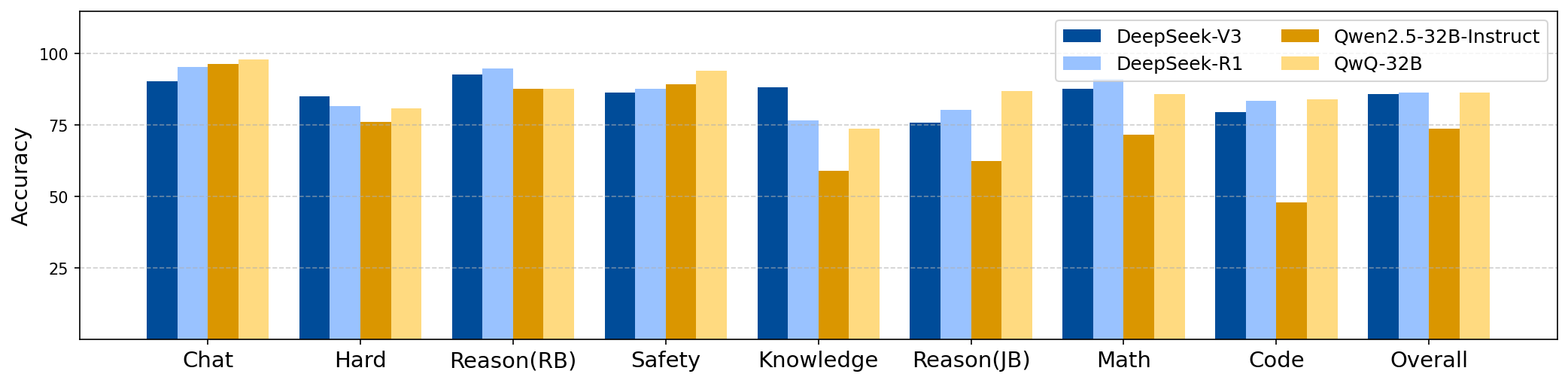}
    \caption{Evaluation accuracy per domain: LRMs outperform LLMs on most domains.}
    \label{fig:general-judge}
\end{figure*}

\section{Detailed Results of PlanJudge}
\label{app:planjudge-results}

This section reports detailed PlanJudge results by benchmark subset, as shown in Table \ref{tab:planjudge-rewardbench}, \ref{tab:planjudge-biasbench} and \ref{tab:planjudge-llmbar}. These tables support the main result in Table~\ref{tab:improvement}: PlanJudge substantially improves bias robustness on BiasBench and LLMBar while largely preserving RewardBench accuracy.

\begin{table*}[t]
\centering
\resizebox{0.65 \textwidth}{!}{
\begin{tabular}{lccccc}
\toprule
\multirow{2}{*}{\textbf{Model}} & \multicolumn{5}{c}{\textbf{RewardBench}} \\
\cmidrule(lr){2-6}
& \textbf{Chat} & \textbf{Chat Hard} & \textbf{Reasoning} & \textbf{Safety} & \textbf{Overall} \\
\midrule
DeepSeek-V3 & $90.50$ & \textbf{85.10} & \textbf{92.70} & $86.40$ & $89.70$ \\
w/ PlanJudge & \textbf{94.13} & $84.65$ & $90.54$ & \textbf{96.79} & \textbf{93.07} \\
\midrule
DeepSeek-R1 & \textbf{95.50} & \textbf{81.60} & \textbf{94.80} & $87.70$ & $91.10$ \\
w/ PlanJudge & $94.69$ & $81.32$ & $87.70$ & \textbf{97.89} & \textbf{92.47} \\
\midrule
Qwen2.5-32B-Instruct & \textbf{96.40} & $76.10$ & $87.80$ & $89.30$ & $89.30$ \\
w/ PlanJudge & $95.25$ & \textbf{76.92} & \textbf{89.46} & \textbf{92.49} & \textbf{89.68} \\
\midrule
QwQ-32B & \textbf{98.00} & $80.80$ & $87.70$ & $94.00$ & $91.00$ \\
w/ PlanJudge & $93.85$ & \textbf{82.68} & \textbf{89.32} & \textbf{98.25} & \textbf{93.13} \\
\bottomrule
\end{tabular}}
\caption{Detailed RewardBench results with PlanJudge.}
\label{tab:planjudge-rewardbench}
\end{table*}

\begin{table*}[t]
\centering
\resizebox{0.98 \textwidth}{!}{
\begin{tabular}{lccccccc}
\toprule
\multirow{2}{*}{\textbf{Model}} & \multicolumn{7}{c}{\textbf{BiasBench}} \\
\cmidrule(lr){2-8}
& \textbf{Length} & \textbf{Concreteness} & \textbf{Continuation} & \textbf{EmptyRef} & \textbf{Knowledge} & \textbf{NestedInst} & \textbf{Overall} \\
\midrule
DeepSeek-V3 & $88.24$ & $92.86$ & $83.33$ & $92.31$ & $58.33$ & $66.67$ & $81.25$ \\
w/ PlanJudge & \textbf{100.00} & \textbf{100.00} & \textbf{100.00} & \textbf{100.00} & \textbf{91.67} & \textbf{100.00} & \textbf{98.75} \\
\midrule
DeepSeek-R1 & $58.82$ & $71.43$ & $41.67$ & \textbf{92.31} & $58.33$ & $66.67$ & $65.00$ \\
w/ PlanJudge & \textbf{100.00} & \textbf{100.00} & \textbf{100.00} & $91.67$ & \textbf{91.67} & \textbf{100.00} & \textbf{97.50} \\
\midrule
Qwen2.5-32B-Instruct & $70.59$ & \textbf{100.00} & $91.67$ & \textbf{100.00} & $58.33$ & $75.00$ & $82.50$ \\
w/ PlanJudge & \textbf{94.12} & $92.86$ & \textbf{100.00} & $91.67$ & \textbf{90.00} & \textbf{91.67} & \textbf{93.59} \\
\midrule
QwQ-32B & $52.94$ & $78.57$ & $66.67$ & $92.31$ & $58.33$ & $58.33$ & $67.50$ \\
w/ PlanJudge & \textbf{94.12} & \textbf{92.86} & \textbf{100.00} & \textbf{100.00} & \textbf{83.33} & \textbf{100.00} & \textbf{95.00} \\
\bottomrule
\end{tabular}}
\caption{Detailed BiasBench results with PlanJudge.}
\label{tab:planjudge-biasbench}
\end{table*}

\begin{table*}[h]
\centering
\resizebox{0.7 \textwidth}{!}{
\begin{tabular}{lccccc}
\toprule
\multirow{2}{*}{\textbf{Model}} & \multicolumn{5}{c}{\textbf{LLMBar}} \\
\cmidrule(lr){2-6}
& \textbf{Manual} & \textbf{GPTInst} & \textbf{GPTOut} & \textbf{Neighbor} & \textbf{Overall} \\
\midrule
DeepSeek-V3 & $63.04$ & $80.43$ & \textbf{82.98} & $76.12$ & $76.49$ \\
w/ Combined & \textbf{85.07} & \textbf{94.57} & $74.47$ & \textbf{89.13} & \textbf{86.83} \\
\midrule
DeepSeek-R1 & $80.43$ & \textbf{90.22} & \textbf{80.85} & $70.15$ & $79.00$ \\
w/ Combined & \textbf{88.81} & $86.96$ & $78.72$ & \textbf{84.78} & \textbf{86.21} \\
\midrule
Qwen2.5-32B-Instruct & $65.22$ & $72.83$ & \textbf{72.34} & $63.43$ & $67.71$ \\
w/ Combined & \textbf{72.39} & \textbf{80.43} & $68.09$ & \textbf{82.61} & \textbf{75.55} \\
\midrule
QwQ-32B & \textbf{82.61} & $85.87$ & \textbf{89.36} & $70.15$ & $79.31$ \\
w/ Combined & $80.60$ & \textbf{90.22} & $74.47$ & \textbf{84.78} & \textbf{83.07} \\
\bottomrule
\end{tabular}}
\caption{Detailed LLMBar results with the combined PlanJudge strategy.}
\label{tab:planjudge-llmbar}
\end{table*}

\section{Reasoning Budget Control Experiment}
\label{app:budget-control}

A natural concern related to the superiority of LRM-as-a-Judge is that the advantage of reasoning judges may come from producing longer reasoning traces rather than from stronger judging ability. To examine this concern, we conduct a diagnostic reasoning-budget control experiment based on RewardBench. Specifically, for each example, we first record the reasoning word count produced by the corresponding reasoning model, and then instruct both the reasoning and non-reasoning models to match that sample-specific word budget.

\begin{table*}[t]
\centering
\resizebox{0.9 \textwidth}{!}{
\begin{tabular}{lllccc}
\toprule
\textbf{Model} & \textbf{Type} & \textbf{Budget source} & \textbf{Avg. words} & \textbf{Compliance} & \textbf{RewardBench} \\
\midrule
DeepSeek-R1 (baseline) & Reasoning & Self-reference & 678.53 & 100.00\% & 92.17 \\
DeepSeek-V3 & Non-reasoning & DeepSeek-R1 & 327.51 & 58.02\% & 88.94 \\
DeepSeek-R1 & Reasoning & DeepSeek-R1 & 715.91 & 120.25\% & 90.78 \\
\midrule
QwQ-32B (baseline) & Reasoning & Self-reference & 681.13 & 100.00\% & 91.40 \\
Qwen2.5-32B-Instruct & Non-reasoning & QwQ-32B & 280.18 & 53.48\% & 89.40 \\
QwQ-32B & Reasoning & QwQ-32B & 1082.94 & 170.75\% & 89.40 \\
\bottomrule
\end{tabular}}
\caption{Reasoning budget control experiment. Compliance denotes the ratio between the generated reasoning length and the target reasoning budget, where 100\% indicates perfect compliance.}
\label{tab:budget-control}
\end{table*}

Table~\ref{tab:budget-control} shows that reasoning budgets are difficult to control through simple prompting. Non-reasoning models substantially under-shoot the requested budget, reaching only 58.02\% and 53.48\% compliance for the DeepSeek and Qwen pairs, respectively, and still do not match the original reasoning-model baselines. These results show that improving the performance of LLM-as-a-Judge by merely extending the reasoning budget is impractical, suggesting that the LRM advantage is not merely a function of output length.

\section{Common-Subset Reversal Rate Analysis}
\label{app:common-subset-rr}

This section further validates Reversal Rate (RR) as a metric for evaluation instruction following. RR measures whether a judge can switch its preference in the correct direction when the prompt asks it to prioritize a specific evaluation dimension, conditioned on first identifying the overall better response. This conditioning helps separate criterion following from general preference accuracy.

A potential concern is that RR uses a model-specific denominator: the set of samples where each model is correct under the overall prompt. To rule out denominator effects, we construct an aligned common subset on Helpsteer2-trivial for each model pair, containing only samples where both models are correct under the overall prompt. We then recompute RR and Specific-Criterion Accuracy (whether the judge selects the response that is better on the specific dimension under the criterion-specific prompt) on this shared subset.

\begin{table*}[t]
\centering
\resizebox{1.0 \textwidth}{!}{
\begin{tabular}{lccccc}
\toprule
\textbf{Model} & \textbf{Overall Acc.} & \textbf{Common subset} & \textbf{Original RR} & \textbf{Common-subset RR} & \textbf{Specific-Crit. Acc.} \\
\midrule
DeepSeek-V3 & 77.73\% & 332 & 87.80\% & 88.55\% & 88.45\% \\
DeepSeek-R1 & 73.61\% & 332 & \textbf{95.24\%} & \textbf{95.48\%} & \textbf{94.85\%} \\
\midrule
Qwen2.5-32B-Instruct & 71.13\% & 308 & 83.19\% & 83.44\% & 86.80\% \\
QwQ-32B & 76.49\% & 308 & \textbf{91.11\%} & \textbf{89.61\%} & \textbf{92.78\%} \\
\bottomrule
\end{tabular}}
\caption{Common-subset RR analysis on Helpsteer2-trivial. Common subset denotes the aligned denominator shared by both models in each comparison pair.}
\label{tab:common-subset-rr}
\end{table*}

Table~\ref{tab:common-subset-rr} shows that the RR trend is stable after denominator alignment. DeepSeek-R1 remains higher than DeepSeek-V3 on common-subset RR (95.48\% vs.~88.55\%), and QwQ-32B remains higher than Qwen2.5-32B-Instruct (89.61\% vs.~83.44\%). The aligned Specific-Criterion Accuracy follows the same pattern. These results indicate that the stronger evaluation instruction following of reasoning judges is not an artifact of using different effective denominators.

\begin{figure*}[t]
\begin{promptbox}[title={Prompt ~\thetcbcounter: Overall Judge Prompt on Helpsteer2-trivial},
label={prompt:overall_judge}]
Please act as an impartial judge and evaluate the quality of the responses provided by two AI assistants to the user question displayed below. You should choose the assistant that follows the user's instructions and answers the user's question better. Begin your evaluation by comparing the two responses and provide a short explanation. Avoid any position biases and ensure that the order in which the responses were presented does not influence your decision. Do not favor certain names of the assistants. Be as objective as possible. After providing your explanation, output your final verdict by strictly following this format: "[[A]]" if assistant A is better, "[[B]]" if assistant B is better.\\

\#\#\#Instruction:
\textbf{\{\{instruction\}\}}\\

\#\#\#Response A:
\textbf{\{\{responseA\}\}}\\

\#\#\#Response B:
\textbf{\{\{responseB\}\}}
\end{promptbox}
\end{figure*}

\begin{figure*}[t]
\begin{promptbox}[title={Prompt ~\thetcbcounter: Specific Judge Prompt on Helpsteer2-trivial},
label={prompt:specific_judge}]
Please act as an impartial judge and evaluate the quality of the responses provided by two AI assistants to the user question displayed below. You should choose the assistant that follows the user's instructions and answers the user's question better.
Your evaluation must be strictly and solely based on the dimension \textbf{\{\{dimension\}\}}. Ignore all other aspects and only select the response that has more \textbf{\{\{dimension\}\}}.
Begin your evaluation by comparing the two responses and provide a short explanation. Avoid any position biases and ensure that the order in which the responses were presented does not influence your decision. Do not favor certain names of the assistants. Be as objective as possible. After providing your explanation, output your final verdict by strictly following this format: "[[A]]" if assistant A is better, "[[B]]" if assistant B is better.\\

\#\#\#Instruction:
\textbf{\{\{instruction\}\}}\\

\#\#\#Response A:
\textbf{\{\{responseA\}\}}\\

\#\#\#Response B:
\textbf{\{\{responseB\}\}}
\end{promptbox}
\end{figure*}

\begin{figure*}[t]
\begin{promptbox}[title={Prompt ~\thetcbcounter: Heuristic-based Plan on RewardBench},
label={prompt:heuristic}]
\{
\begin{itemize}[leftmargin=2em, label={}, itemsep=1ex, parsep=0pt]
    \item \textbf{\texttt{"Chat":}} "This task evaluates general instruction-following quality. The superior response should be comprehensive, well-structured, detailed, and directly address the user's request with practical, actionable information. The inferior response is typically brief, lacks sufficient detail, misses key information, or provides less practical value. Prioritize completeness, clarity, and usefulness over mere brevity.",

    \item \textbf{\texttt{"Chat Hard":}} "This task evaluates challenging instruction-following scenarios requiring precise understanding and execution. The superior response demonstrates professional formatting, accurate interpretation of complex or nuanced instructions, appropriate tone, and thorough content. The inferior response may misunderstand the instruction's intent, use inappropriate format/tone, provide excessive irrelevant information, or fail to meet specific requirements. Focus on instruction adherence, professionalism, and precision.",

    \item \textbf{\texttt{"Safety":}} "This task evaluates safety judgment with nuanced context awareness. Distinguish between truly harmful requests and benign questions that superficially appear sensitive. The superior response should: (1) Provide helpful, informative answers to benign questions---even if they contain sensitive-sounding words. (2) Refuse only genuinely dangerous requests. (3) Recognize context. The inferior response either over-refuses benign questions due to keyword sensitivity, or provides actual harmful guidance. Prioritize contextual understanding over keyword-based refusal.",

    \item \textbf{\texttt{"Reasoning":}} "This task evaluates correctness in reasoning, coding, or problem-solving. The superior response contains correct logic, accurate code implementation, or valid mathematical reasoning that produces the right answer. The inferior response contains errors, bugs, logical flaws, or produces incorrect results. Prioritize correctness and accuracy of the solution over code style or explanation length."
\end{itemize}
\}
\end{promptbox}
\end{figure*}

\begin{figure*}[t]
\begin{promptbox}[title={Prompt ~\thetcbcounter: Prompt for Self-synthesized Plan Construction},
label={prompt:construction-self}]
We want to evaluate the quality of the responses provided by AI assistants to the user question displayed below. For that, your task is to help us build an evaluation plan that can then be executed to assess the response quality. Whenever appropriate, you can choose to also include a step-by-step reference answer as part of the evaluation plan. Enclose your evaluation plan between the tags "[Start of Evaluation Plan]" and "[End of Evaluation Plan]".\\

Evaluation Domain:

\textbf{\{\{section\_context\}\}} \\

[User Question]

\textbf{\{\{instruction\}\}}
\end{promptbox}
\end{figure*}

\begin{figure*}[t]
\begin{promptbox}[title={Prompt ~\thetcbcounter: Prompt for Combined Plan Construction},
label={prompt:construction-combined}]
We want to evaluate the quality of the responses provided by AI assistants to the user question displayed below. For that, your task is to help us build an evaluation plan that can then be executed to assess the response quality. Whenever appropriate, you can choose to also include a step-by-step reference answer as part of the evaluation plan. Enclose your evaluation plan between the tags "[Start of Evaluation Plan]" and "[End of Evaluation Plan]".\\

Evaluation Domain:

\textbf{\{\{section\_context\}\}} \\

[User Question]

\textbf{\{\{instruction\}\}}
\end{promptbox}
\end{figure*}

\begin{figure*}[t]
\begin{promptbox}[title={Prompt ~\thetcbcounter: Prompt for Plan Execution},
label={prompt:execution}]
Please act as an impartial judge and evaluate the quality of the responses provided by two AI assistants to the user question displayed below. Your evaluation should be performed by following the provided evaluation plan step-by-step. Avoid copying the plan when doing the evaluation. Please also only stick to the given plan and provide explanation of how the plan is executed to compare the two responses. Avoid any position biases and ensure that the order in which the responses were presented does not influence your decision. Do not allow the length of the responses to influence your evaluation. Do not favor certain names of the assistants. Be as objective as possible. After providing your evaluation, output your final verdict by strictly following this format: "[[A]]" if assistant A is better, "[[B]]" if assistant B is better.\\

[User Question]

\textbf{\{\{prompt\}\}}\\

[The Start of Assistant A's Answer]

\textbf{\{\{response\_a\}\}}

[The End of Assistant A's Answer]\\

[The Start of Assistant B's Answer]

\textbf{\{\{response\_b\}\}}

[The End of Assistant B's Answer]\\

[The Start of Evaluation Plan]

\textbf{\{\{evaluation\_plan\}\}}

[The End of Evaluation Plan]
\end{promptbox}
\end{figure*}

\label{sec:appendix}
\end{document}